\pdfoutput=1
\documentclass[12pt]{article}
\usepackage{mathrsfs,verbatim}
\usepackage{appendix}
\usepackage{natbib}
\usepackage{extarrows}
\usepackage{graphicx,setspace,lscape,longtable,epstopdf,xr}
\usepackage{mathrsfs,amsmath,amsthm,amssymb,color}
\usepackage{epsfig,graphicx}
\usepackage{rotating}
\usepackage{float}
\usepackage{bm}
\usepackage{ragged2e}
\usepackage{todonotes}
\usepackage{hyperref}
\usepackage{multirow} 
\usepackage{makecell}
\usepackage[linesnumbered,ruled]{algorithm2e}
\usepackage{setspace}
\graphicspath{{figure/}}

\hypersetup{
colorlinks=true,
linkcolor=blue,
citecolor=blue,
urlcolor=blue}
\usepackage{booktabs}
\usepackage{algpseudocode}
\usepackage{amsmath}
\usepackage{graphicx}
\usepackage{subfigure}
\usepackage{hyperref}

\bibpunct{(}{)}{;}{a}{,}{,}

\newtheorem{theorem}{Theorem}
\newtheorem{lemma}{Lemma}
\newtheorem{proposition}{Proposition}

\newcommand{\csection}[1]
{\begin{center}
\stepcounter{section}
{\bf\large\arabic{section}. #1}
\end{center}
}

\newcommand{\csubsection}[1]{
\begin{center}
\stepcounter{subsection}
{\it\arabic{section}.\arabic{subsection}. #1}
\end{center}
}

\def\beq{\begin{equation}}
\def\eeq{\end{equation}}
\def\beqr{\begin{eqnarray}}
\def\eeqr{\end{eqnarray}}
\def\beqrs{\begin{eqnarray*}}
\def\eeqrs{\end{eqnarray*}}
\def\bet{\begin{theorem}}
\def\eet{\end{theorem}}
\def\bel{\begin{lemma}}
\def\eel{\end{lemma}}
\def\bep{\begin{proposition}}
\def\eep{\end{proposition}}
\def\bg{\begin{figure}[tbph]\begin{center}}
\def\eg{\end{center}\end{figure}}

\def\bc{\begin{center}}
\def\ec{\end{center}}

\def\mD{\mathcal D}

\def\mR{\mathbb{R}}

\def\mS{\mathbb S}

\def\mF{\mathcal F}

\def\mS{\mathcal S}

\def\argmax{\mbox{argmax}}

\newcommand{\RNum}[1]{\uppercase\expandafter{\romannumeral #1\relax}}

\def\bg{\mbox{\boldmath $g$}}

\renewcommand{\arraystretch}{1.3}

\numberwithin{equation}{section}

\def\be{\begin{eqnarray*}}
\def\ese{\end{eqnarray*}}
\def\be{\begin{eqnarray}}
\def\ee{\end{eqnarray}}
\def\bsq{\begin{equation*}}
\def\esq{\end{equation*}}
\def\bq{\begin{equation}}
\def\eq{\end{equation}}

\def\mR{\mathbb{R}}

\def\argmax{\mbox{argmax}}

\def\mS{\mbox{ $\mathcal{S}$}}

\def\bg{\boldsymbol\gamma}

\def\0{{\bf 0}}
\def\1{{\bf 1}}

\def\bq{\begin{equation}}
\def\eq{\end{equation}}

\def\squarebox#1{\hbox to #1{\hfill\vbox to #1{\vfill}}}

\def\mF{\mathcal{F}}

\def\bse{\begin{eqnarray*} }
\def\ese{\end{eqnarray*}}
\def\be{\begin{eqnarray}}
\def\ee{\end{eqnarray}}
\def\bsq{\begin{equation*}}
\def\esq{\end{equation*}}
\def\bq{\begin{equation}}
\def\eq{\end{equation}}

\def\boxit#1{\vbox{\hrule\hbox{\vrule\kern6pt\vbox{\kern6pt#1\kern6pt}\kern6pt\vrule}\hrule}}

\begin{document}
\begin{center}
{\bf\Large Embedding Compression for Text Classification Using Dictionary Screening}\\
\bigskip

 Jing Zhou$^1$, Xinru Jing$^1$, Muyu Liu$^1$, and Hansheng Wang$^2$
\\

{\it $^1$Center for Applied Statistics and School of Statistics, Renmin University of China \\ $^2$Guanghua School of Management, Peking University}

\end{center}

\begin{singlespace}
\begin{abstract}

In this paper, we propose a dictionary screening method for embedding compression in text classification tasks. The key purpose of this method is to evaluate the importance of each keyword in the dictionary. To this end, we first train a pre-specified recurrent neural network-based model using a full dictionary. This leads to a benchmark model, which we then use to obtain the predicted class probabilities for each sample in a dataset. Next, to evaluate the impact of each keyword in affecting the predicted class probabilities, we develop a novel method for assessing the importance of each keyword in a dictionary. Consequently, each keyword can be screened, and only the most important keywords are reserved. With these screened keywords, a new dictionary with a considerably reduced size can be constructed. Accordingly, the original text sequence can be substantially compressed. The proposed method leads
to significant reductions in terms of parameters, average text sequence, and dictionary size. Meanwhile, the prediction power remains very competitive compared to the benchmark model. Extensive numerical studies are presented to demonstrate the
empirical performance of the proposed method. \\


\end{abstract}
\end{singlespace}

\newpage

\csection{INTRODUCTION}

Over the past few decades, natural language processing (NLP) has become a popular research field. Among the applications of this filed, text classification is considered to be a problem of great importance. Many successful applications exist, such as news classification \citep{2019Revisiting}, topic labeling \citep{auer2007dbpedia}, sentiment analysis \citep{2020SentiLSTM}, and many others. Unlike structured data, natural language comprises typical, unstructured text data, which are very challenging to model. To tackle these challenges, many algorithms and methods have been developed. These methods include, for example, the $N$-gram model \citep{1980Interpolated,Improved1995}, convolutional neural network (CNN)-based models \citep{Kim2014,Zhao2015}, recurrent neural network (RNN)-based models \citep{1997Long,2014Learning}, attention-based models \citep{2014Attention}, transformer-based models \citep{devlin2019bert}, and many others. Among these methods, several RNN-based models and their variants are widely used and have been well-studied. However, applications of these RNN-based models usually incorporate a large number of parameters and thus require substantial memory support. This makes it difficult to deploy these models for embedded systems (e.g., mobile devices and tablets), which are limited in memory, computing power, and battery life. Therefore, models with comparable prediction performances but much reduced sizes are practically needed. This makes model compression a pressing problem.

Note that, for most text classification tasks, the inputs are documents constructed from word sequences. 
Therefore, a standard RNN-based model can be readily applied. Remarkably, the model complexity of an RNN-based model is mainly determined by two factors. The first factor is the model structure, which includes the number of hidden layers, number of neurons in each hidden layer, and dimension of the embedding space. The second important factor is dictionary size. Obviously, large dictionaries lead to more complicated RNN models with a large number of parameters. To illustrate this idea, consider, for example, a standard RNN model with one embedding layer and one recurrent hidden layer. The total number of parameters needed is $dk+2k^2$, where $d$ is the dictionary size and $k$ is the dimension for both the hidden layer and embedding space. 
In the AG's News dataset \citep{2015Zhang}, for instance, the total number of keywords contained in the dictionary could be as large as $d=93,994$. If, following \cite{2014Cho}, we set the dimensions of both the hidden layer and embedding space to $k=128$, then the total number of parameters is more than 12 million. 
As to be demonstrated later, we find that the dictionary size can be effectively reduced to be $d = 3,000$, if the method developed in this work is used. As a result, the total number of parameters can be reduced to about 0.58 million. This accounts for only about 4.76\% of the original model complexity with limited sacrifice of prediction accuracy. 

We are thus motivated to fill the above important theoretical gap by developing a novel dictionary screening method. Dictionary screening excludes less useful keywords from a dictionary. It can be used to effectively reduce dictionary size, leading to a significant reduction in model complexity. 
Specifically, we develop here a novel dictionary screening method as follows. First, we train a pre-specified RNN-based model using the full dictionary on a training dataset. This leads to a benchmark model. With the help of the benchmark model, we obtain the predicted class probabilities for every sample in the validation set. Next, for a given keyword in the dictionary, we consider whether it should be excluded from the dictionary. Obviously, keywords that significantly impact the predicted class probabilities should be kept, while those that do not should be excluded. Thus, the key here is determining how to evaluate the impact of the target keyword in affecting the estimated class probabilities. 

To this end, for each document, we replace the target keyword with a meaningless substitute, which is often an empty space. By doing so, the input of the target keyword is excluded. We then apply the pretrained benchmark model to this new document. This leads to a new set of estimated class probabilities. Thereafter, for each document, we obtain two sets of estimated class probabilities. One is computed based on the full dictionary, and the other is computed based on the dictionary with the target keyword excluded. Next, the difference between the two sets of probability estimators is evaluated and summarized. Each keyword in the dictionary should be evaluated. Keywords with large differences in class probability estimators should be kept. In contrast, those with small differences should be excluded. By selecting an appropriate threshold value, a new dictionary with a substantially compressed size can be obtained. Finally, with the compressed dictionary, each document can be re-constructed. The associated RNN-based model can be re-trained on the re-constructed documents, and its prediction accuracy on the validation dataset can be evaluated. Our extensive numerical experiments suggest that the proposed method can compress the parameter quantity by more than 90\%, on average, with little accuracy sacrificed. 

The main contribution of our work is the development of a compression method for text classification using dictionary screening. There has been relatively little work on compressing dictionary size in the previous literature. A second contribution is that we provide a novel method for evaluating the importance of the keywords in a dictionary. We empirically show that our method outperforms popular baselines like term frequency-inverse document frequency (TF-IDF) and the $t$-test for keyword importance analysis. 
The rest of the article is organized as follows. Section 2 reviews some of
the related literature. Section 3 develops the dictionary screening method. Section 4 presents extensive experiments on the proposed method, and the results are summarized in Section 5. Section 6 concludes the paper with a brief discussion on future research.

\csection{RELATED WORKS}

In this section, we first review some classical compression methods in the deep learning field, and then we focus on neural model compression in the
context of text classification. Following the review paper by \citet{deng2020model}, the current model compression methods for RNN-based models can be categorized into the following four categories: model compacting, tensor decomposition, pruning, and data quantization. By model compacting, we mean that unimportant network modules in the original model structure should be merged or removed. By doing so, a new and simplified model can be obtained. Typical examples are the GRU model \citep{chung2014empirical}, S-LSTM \citep{wu2016investigating}, and JANET model \citep{van2018unreasonable}. Through tensor decomposition, researchers aim to replace the original weight matrix with its low rank approximations, such as full-rank decomposition \citep{sainath2013low, liu2015l_}, singular value decomposition \citep{li2007neural, xue2014singular},
QR decomposition \citep{yu2018gradiveq, aizenberg2012modified}, and CUR decomposition \citep{thurau2012deterministic, gittens2016revisiting}. Recently, a progressive principal component analysis method has also proved useful \citep{ZHOU2021,Qi2022}. The pruning method generally aims at pruning the unimportant parameters in networks. Typical examples are weight pruning \citep{2017Luo}, channel pruning \citep{2018Hu}, kernel pruning \citep{2016PLi}, and neuron pruning \citep{2016Hu}. Data quantization refers to methods that transform the floating-point operation of a neutral network into a fix-point operation \citep{xu2018alternating,wang2018hitnet}.

In the context of text classification, most of the prior work focuses on compressing the embedding matrix using different methods. For example, a number of researchers have adopted hashing or quantization-based approaches to compress the embedding matrix \citep{2016Joulin,2017Raunak,2017Shu}. \cite{acharya2019online} proposed a low rank matrix factorization for the embedding layer. In addition to compressing the embedding matrix, there is another branch of research that shows training with character-level inputs can achieve
several benefits over word-level approaches, and it does so with fewer parameters \citep{2016Xiao}. Despite the excellent research that has been done on model compressing, it seems that most studies focus on simplifying the model structure in one way or another. Little research has been done on dictionary screening. As discussed in the introduction, we can see that the size of the dictionary can have an important impact on model size. As a consequence, we are motivated to fill this gap by proposing a dictionary screening method for text classification applications.

\csection{METHODOLOGY}

\csubsection{Problem set up}

Let $\mD = \{w_{d}: 0\leq d \leq D\}$ be a dictionary containing a total of $D$ keywords with $w_{d}$ representing the $d$th keyword. We define $w_{0}$ to be an empty space. Then, assume a total of $N$ documents indexed by $1\leq i \leq N$. Let $X_{i} = \{X_{it}\in\mD: 1\leq t\leq T\}$ be the $i$th document. The document is constructed by a sequence of keywords, $X_{it}$, which is indexed by $t$ and is generated from $\mD$. If the actual document length, $T^{*}$, is less than $T$, we then define $X_{it} = w_{0}$ for $T^{*}<t\leq T$. Next, let $Y_{i}\in \{1,2,\cdots,K\}$ be the class label associated with the $i$th document. The goal is then to train a classifier so that we can accurately predict $Y_{i}$. To this end, various deep learning models can be used. For illustration purposes, we consider here a simple RNN model with four layers: one input layer of a word sequence with dimension $T$, one embedding layer with $d_{1}=128$ hidden nodes, one simple RNN layer with $d_{2}=64$ hidden nodes, and one fully connected layer with $K$ nodes (i.e., the number of class labels). Suppose the dictionary size is $D=10,000$ and the number of class labels is $K=10$, then for the above simple RNN model, the total number of parameters is given by $df_{a} =  10,000 \times 128+64 \times 128+64 \times 64+64\times10 = 1,292,928$ with the bias term ignored. However, this number will be much reduced if the dictionary size can be significantly decreased. For example, the total number of parameters will be reduced to $df_{b}= 1,000 \times128+64\times128+64\times64+64\times10 = 140,928 $ if $D=1,000$. This represents a model complexity reduction as large as $(1-df_{b}/df_{a})\times 100\% = 89.1\%$. We are then motivated to develop a method for dictionary screening.

\csubsection{Dictionary screening} 

As discussed in the introduction, one of the key tasks for dictionary screening is to evaluate the importance of each keyword in the dictionary. We can formulate the problem as follows. First, we train a pre-specified RNN-based model using the full dictionary, $\mD$, on the training dataset. Mathematically, we can write this model as $f(X_{i},\theta) = \{f_{k}(X_{i},\theta)\}\in\mR^{K}$, where $X_{i}$ is the input document, with $\theta$ as the unknown parameters that need to be estimated. Note that $f(X_{i},\theta)$ is a $K$-dimensional vector, its $k$th element, $f_{k}(X_{i},\theta)\in[0,1]$, is a theoretical assumed function to approximate the class probability. That is, $P(Y_{i}=k | X_{i})\approx f_{k}(X_{i},\theta)$. By the universal approximation theorem \citep{1989Cy,1989Multilayer}, we know that this approximation can be arbitrarily accurate as long as the approximation function, $f(\cdot,\theta)$, can be sufficiently flexible. To estimate $\theta$, an approximately defined loss function (e.g., the categorical cross entropy) is usually used. Denote the loss function as $\mathcal{L}_{N}(\theta) = N^{-1}\sum_{i=1}^{N}\ell(X_{i},\theta)$, where $\ell(X_{i},\theta)$ is the loss function evaluated on the $i$th document. The parameter estimators can then be obtained as $\hat\theta = \argmax{\mathcal{L}_{N}(\theta)}$. This leads to the pretrained model as $f(X_{i},\hat\theta)$, which serves as the benchmark model. 

Next, with the help of the pretrained model, we consider how to evaluate the importance of each keyword in $\mD$. Specifically, consider the $d$th keyword, $w_{d}$, in $\mD$ with $1\leq d \leq D$.
Define $\mS_{\mF}=\{1,2,\cdots,N\}$ as the indices for the full training document. 
Then, for every $i\in\mS_{\mF}$, we compute its estimated class probability vector as $\widehat p_{i}= f(X_{i},\hat\theta)$. Next, for the document $X_{i}=\{X_{it}\in\mD:1\leq t \leq T\}$, we generate a document copy as $X_{i}^{(d)} =\{ X_{it}^{(d)}\in\mD: 1\leq t \leq T\}$, where $X_{it}^{(d)} = X_{it}$ if $X_{it} \neq w_{d}$, and $X_{it}^{(d)} = w_{0}$ if $X_{it} = w_{d}$. In other words, $X_{i}^{(d)}$ is a document that is almost the same as $X_{i}$. The only difference is that keyword $w_{d}$ is replaced by an empty space, $w_{0}$. We next apply the benchmark model to $X_{i}^{(d)}$, so an update class probability vector, $\widehat p_{i}^{(d)}= f(X_{i}^{(d)},\hat\theta)$, can be obtained. The difference between $\widehat p_{i}$ and $\widehat p_{i}^{(d)}$ is evaluated by their $\ell_{2}$-distance as $|| \widehat p_{i} - \widehat p_{i}^{(d)}||^2$. We then summarize the difference for every $w_{d} \in \mD$ as $\widehat \lambda(d) = |\mS_{\mF}|^{-1}\sum_{i\in\mS_{\mF}}|| \widehat p_{i} - \widehat p_{i}^{(d)}||^2$, where $|\mS_{\mF}|$ is the size of $\mS_{\mF}$. This is further treated as the important score for each keyword, $w_{d} \in \mD$. Because this important score is obtained by evaluating the differences in class probability estimators, we name it as the CPE method for simplicity.

Finally, with a carefully selected threshold value, $\lambda$, a new dictionary can be constructed as $\mD_{\lambda} = \{w_{d}\in \mD: \widehat\lambda(d)\geq \lambda\}\cup\{w_{0}\}$. To this end, each document $X_{i}$ can be reconstructed as $X_{\lambda_{i}} = \{X_{\lambda_{i}t}\in\mD_{\lambda}:1\leq t\leq T\}$, where $X_{\lambda_{i}t}=X_{it}$ if $X_{it}\in \mD_{\lambda}$, and $X_{\lambda_{i}t}=w_{0}$ otherwise. Then, by replacing $X_{i}$s in the loss function with $X_{\lambda_{i}}$, a new set of parameter estimators can be obtained as $\widehat\theta_{\lambda}= \argmax{\mathcal{L}_{\lambda}(\theta)}$, where ${\mathcal{L}_{\lambda}(\theta)} = N^{-1}\sum_{i=1}^{N}\ell(X_{\lambda_{i}},\theta)$. Once $\widehat\theta_{\lambda}$ is obtained, the prediction accuracy of the resulting model, $f(X_{\lambda_{i}},\widehat\theta_{\lambda})$, can be evaluated on the testing dataset. Thereafter, the resulting model, $f(X_{\lambda_{i}},\widehat\theta_{\lambda})$, serves as the reduced model after applying dictionary screening.
The algorithm details are presented as follows.

\begin{algorithm}
\setstretch{1}

    \SetKwInOut{Input}{Input}
    \SetKwInOut{Output}{Output}
    \DontPrintSemicolon
    
    \SetKwFunction{Bf}{Train a benchmark model}
    \SetKwFunction{Ef}{Evaluate the importance of each keyword in $\mathcal{D}$}
    
    \SetKwProg{Fn}{Procedure1}{:}{}
    \Fn{\Bf}{
        \Input{Document $X_{i}=\{X_{it}\in\mD:1\leq t \leq T\}$; $Y_{i}\in \{1,2,\cdots,K\}$ for $1 \le i \le N$}
        \Output{Benchmark model $f(X_i,\hat \theta)$ with estimated parameters $\hat \theta$ for $1 \le i \le N$}
        $\hat \theta = argmax \mathcal{L}_N(\theta)$\;
        with $\mathcal{L}_N(\theta) =  {N} ^{-1} \sum\nolimits_{i \in 1}^N{\ell(X_i,  \theta)}$\;
    where$ \ell(X_i,\theta)$ is the loss evaluated on the $i$th document\;
    }
    \SetKwProg{Fn}{Procedure2}{:}{}
    \Fn{\Ef}{
        \Input{$\mathcal{D}= \{ w_d: 0 \le d \le D \}$ , $f(X_i,\hat \theta)$ for $1 \le i \le N$}
        \Output{$\mD_{\lambda} = \{w_{d}\in \mD: \widehat\lambda(d)\geq \lambda\}\cup\{w_{0}\}$}
        \For{$d = 1$ to $D$}{
        \For{$i\in\mS_{\mF}$,$\mS_{\mF}=\{1,2,\cdots,N\}$ is the indices for the full training document.}
        {
          $X_{i}=\{X_{it}\in\mD:1\leq t \leq T\}$\;
          Generate a copy as $X_{i}^{(d)} =\{ X_{it}^{(d)}\in\mD: 1\leq t \leq T\}$
          following:\\
          \eIf{$X_{it} \neq w_{d}$}
          {
          $X_{it}^{(d)} = X_{it}$\;
          }
          {
          $X_{it}^{(d)} = w_{0}$\;
          }
Uncompressed class probability : $\widehat p_{i}= f(X_{i},\hat\theta)$  \\
Compressed class probability : $\widehat p_{i}^{(d)}= f(X_{i}^{(d)},\hat\theta)$ \\
Difference computed by $\ell_{2}$-distance as $|| \widehat p_{i} - \widehat p_{i}^{(d)}||^2$\;}
Summarize the difference:\;
$\widehat \lambda(d) = |\mS_{\mF}|^{-1}\sum_{i\in\mS_{\mF}}|| \widehat p_{i} - \widehat p_{i}^{(d)}||^2$ where $|\mS_{\mF}|$ is the size of $\mS_{\mF}$\;
}
Select a threshold value $\lambda$
to finally obtain $\mD_{\lambda} = \{w_{d}\in \mD: \widehat\lambda(d)\geq \lambda\}\cup\{w_{0}\}$
}
\caption{Dictionary screening method}   
\end{algorithm}
 
\csection{EXPERIMENTS}

\csubsection{Task Description and Datasets}

To demonstrate its empirical performance, the proposed dictionary screening method is evaluated on four large-scale datasets covering various text classification tasks. These are, respectively, news classification (AG's News and Sougou News), sentiment analysis (Amazon Review Polarity, ARP), and entity classification (DBPedia). These datasets are popularly studied in previous literature \citep{2015Zhang,2016Xiao}. Summary statistics of the four large-scale datasets are presented in Table \ref{dataset}. For all the datasets (except for Sougou News), the sample size of each category is equal in both the training and testing sets. Take AG's News for example, it has 30,000 samples and 1,900 samples per class in the training set and testing set, respectively. For more detailed information about the four datasets, see \cite{2015Zhang}. It should be noted that, to make the experiments more diverse, the Sougou News data used in this paper are different from those in \cite{2015Zhang}. Particularly, we used the original Chinese characters of Sougou News to test the proposed method, not the Pinyin style used in \cite{2015Zhang}'s work. Moreover, unlike the other three datasets, the sample size of each category (e.g., sports, entertainment, business, and the Internet) in Sougou News is not equal. The proportions of the four categories in the training and testing sets are 48\%, 15\%, 25\%, and 12\%, respectively. 

\begin{table}[htp]
\caption{Summary of four large-scale datasets.}\label{dataset}
\vspace{0.25cm}
\renewcommand\arraystretch{1.2}
\centering
\begin{tabular}{ccccc}
\hline\hline
Dataset&Classes&Task&TrainingSize&TestingSize\\ \hline
AG's News       &4 &news classification &120,000 &7,600 \\
Sougou News      &4 &news classification  &63,146 &15,787\\
DBPedia &14 &entity classification &560,000 &70,000 \\
Amazon Review Polarity &2&sentiment analysis& 3,600,000 &400,000 \\

\hline\hline
\end{tabular}
\end{table}

\csubsection{Model Settings}

We consider here two different types of deep learning models for text classification. They are, respectively, TextCNN \citep{2014Kim} and TextBiLSTM \citep{2016Li}.
We follow their network structures but with some modifications to adapt to our experiments. In the task of text classification, the input is text sequence $X_{i}$ with length $T$. It should be noted that $T$ is different for different datasets. In the current experiment, to train the benchmark models, $T$ is set to 60, 300, 50, and 100 for AG's News, Sougou News, DBPedia, and Amazon Review Polarity, respectively. Practically, each keyword $w_{d}\in\mD$ in the text sequence will be converted to a high dimensional vector of $d_{1}$ via an embedding layer \citep{2013Efficient}. For all three models, the embedding size, $d_{1}$, is set to 128. Next, we briefly describe the construction details for the two models. 

{\it TextCNN.} After the embedding layer, we use three convolutional layers to extract text information. Each convolutional layer has $d_{1}=128$ filters with kernel size $k\in\{3,4,5\}$, followed by a max pooling with receptive field size $r=1$. Rectified linear units (ReLUs) \citep{2011Deep} are used as activation functions in the convolutional layers. Then, we concatenate the max pooling results of the three layers and pass it to the final dense layer through a softmax function for classification.

\begin{figure}[ht]{}
  \centering{}
  \includegraphics[width=1.0\textwidth]{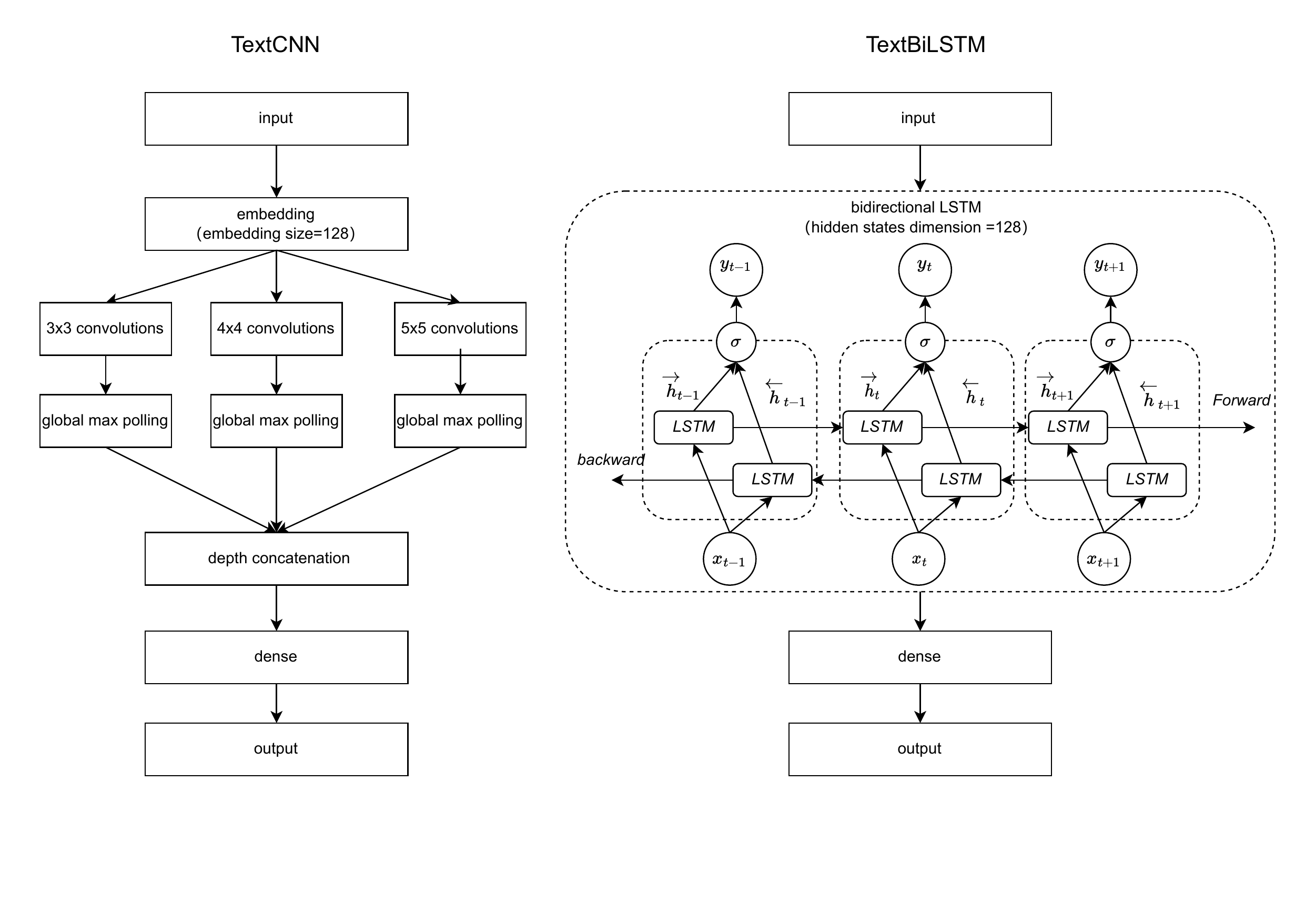}
  \caption{A graphical illustration of model structures for TextCNN and TextBiLSTM.}
  \label{model structure}
\end{figure}

{\it TextBiLSTM.} The bidirectional LSTM (bi-LSTM) can be seen as an improved version of the LSTM. This model structure can consider not only forward encoded information, but also reverse encoded information \citep{2016Li}. We apply a bi-LSTM layer with the hidden states dimension, $d_{2}$, set to 128. We then use the representation obtained from the final timestep (e.g., $X_{iT}$) of the bi-LSTM layer and pass it through a softmax function for text classification. A graphical illustration of the two model structures is presented in Figure \ref{model structure}. 

\csubsection{Tuning Parameter Specification} 

The implementation of the proposed dictionary screening method involves a tuning parameter, which is the threshold value $\lambda$. For a given classification model, this tuning parameter should be carefully selected to achieve the best empirical performance. Here, the best empirical performance means that the proposed dictionary screening method can reduce the number of model parameters as much as possible under the condition of ensuring little or no loss of accuracy. Generally, the larger the $\lambda$ value is, the smaller the size of the screened dictionary, and thus the higher the reduction rate that can be achieved. However, considerable prediction accuracy might be lost. In our experiments, different $\lambda$ values indicate that different numbers of keywords can be reserved for subsequent text compression. To this end, for analysis simplicity, we investigate the number of important keywords reserved, denoted as $K$. Specifically, we rearrange the keywords in descending order according to the importance score (e.g., $\widehat \lambda(d)$), and we select the top 1000, 3000, and 5000 keywords, respectively. That is, $K = \{1000,3000,5000\}$. Therefore, we can evaluate the impact of different tuning parameters on the performance of the proposed dictionary screening method.

\csubsection{Competing Methods}

For comparison purposes, two other methods for evaluating the importance of the keywords in a dictionary are studied. The first one is to calculate the TF-IDF \citep{1972tfidf} value of each keyword $w_{d}\in\mD$. For the $k$th keyword, we use the word counts as the term-frequency (TF). The inverse document frequency (IDF) is the logarithm of the division between the total number of documents and the number of documents with the $k$th word in the whole dataset. To this end, the TF-IDF value for each $w_{d}\in\mD$ can be obtained by multiplying the values of TF and IDF. It is remarkable that the larger the TF-IDF value is, the more important the keyword is. The second method is to compute a $t$-test type statistic. Recall that, for the $d$th keyword, $w_{d}\in\mD$, we have two sets of class probabilities, $\widehat p_{i}$ and $\widehat p_{i}^{(d)}$. Both are $K$-dimensional vectors. Then, for each dimension $k\in K$, a standard paired two sample $t$-test can be constructed to test for the statistical significance. The resulting $p$ values obtained from
different $k$s (e.g., different categories) are then summarized, and the smallest one is selected as the final $t$-test type measure for the target keyword, denoted as $P_{i,d}$. In this case, the smaller the $P_{i,d}$ value is, the more important the keyword is. 

In summary, we have three methods to evaluate the importance of each keyword in $\mD$. These are the proposed method for evaluating the differences in class probability estimators (CPE), the method for evaluating the TF-IDF values (TF-IDF), and the method for evaluating the $t$-test type statistics ($t$-statistic). To make a fair comparison, the new dictionaries constructed by the three methods are of equal size (e.g., with same tuning parameter $K$). Then, following the procedure described in Section 3.2, we can obtain three different prediction accuracies based on the screened dictionary. 

\csubsection{Performance Measures and Implementation}

Following the existing literature, \citep{2015Zhang,2016Xiao,acharya2019online}, and our own concerns, we adopt four measures to gauge the empirical performances of the different compression
methods: the parameter reduction ratio (Prr), dictionary reduction ratio (Drr), 
and reduction ratio for averaged text sequence (Trr). Meanwhile, the out-of-sample prediction accuracy (Acc) is also monitored.

Both text classification models (e.g., TextCNN and TextBiLSTM) are trained on the four large-scale datasets. This leads to a total of $2\times 4 = 8$ working models. All the working models are trained using the AdaDelta (Zeiler, 2012) with $\rho = 0.95, \epsilon = 10^{-5}$, and a batch size of 128. The weight decay is set to $5\times 10^{-4}$ with an $\ell_{2}$-norm regularizer. To prevent overfitting, the dropout and early stopping strategies are used for different working models. Finally, a total of 200 epochs are conducted for each working model. For each working model, we choose the epoch with the maximum prediction accuracy on the validation set as the baseline model. All the experiments were run on a Tesla P100 GPU with 64 GB memory.

\csection{RESULTS ANALYSIS}

\csubsection{Tuning Parameter Effects}

\begin{figure*}[htb]
\centering
\includegraphics[width=15cm]{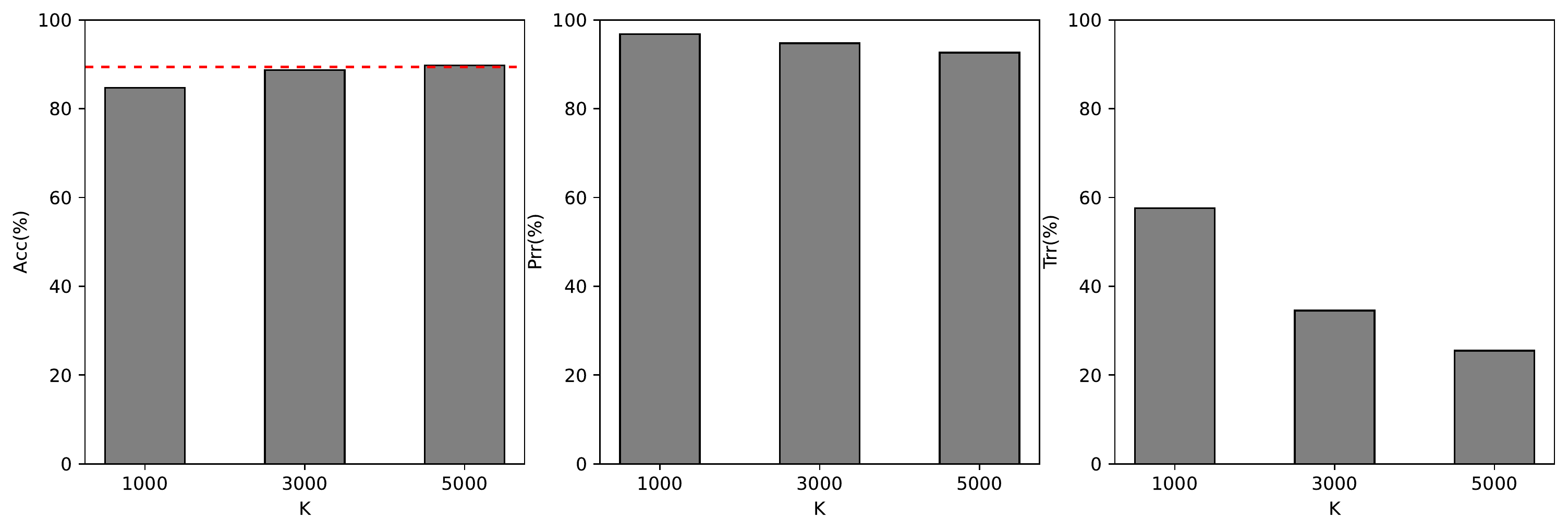}
\includegraphics[width=15cm]{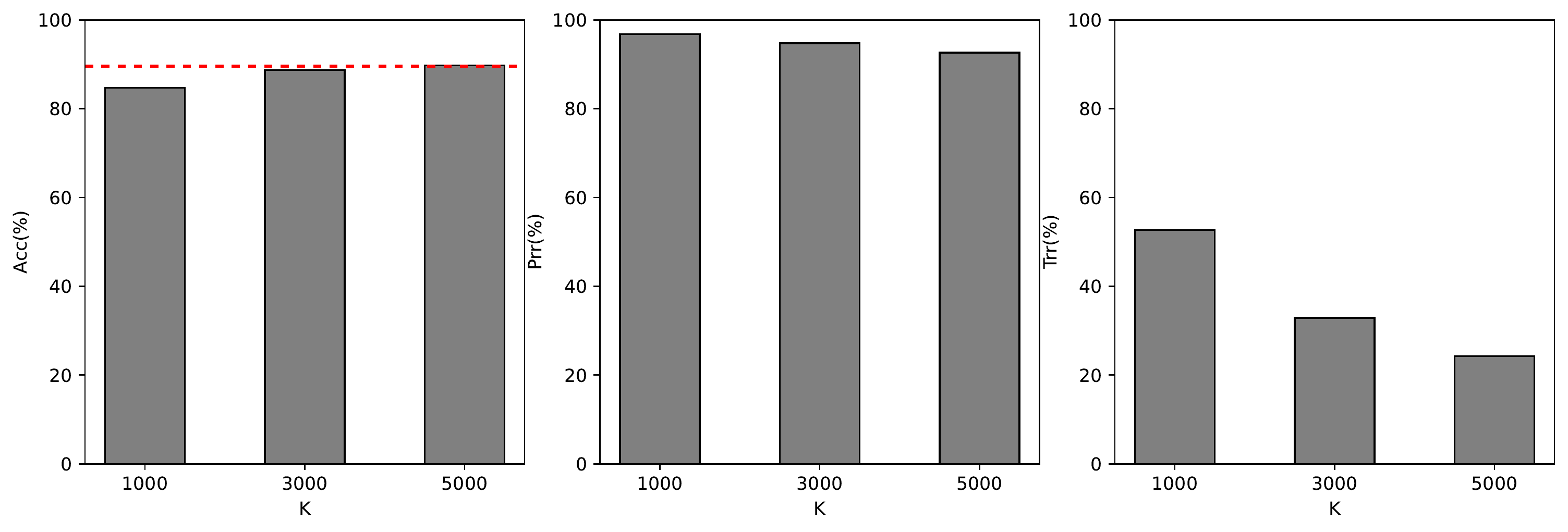}
\caption{Detailed experimental results for TextCNN and TextBiLSTM on the AG's News dataset. Three different $K$ values are considered ($K=$ 1000,3000,5000). Three performance measures are
summarized: prediction accuracy (Acc), parameter reduction ratio (Prr), and reduction ratio for averaged text sequence (Trr). The red dashed line in the left panel represents the accuracy of the
benchmark model. }\label{tuning}
\end{figure*}

In this subsection, we study the impact of the tuning
parameter, $K$, which determines the number of keywords reserved. Three measures are used to gauge the finite sample performance: Acc, Prr, and Trr. For illustration purposes, we use the AG's News dataset as an example. For this experiment, three different $K$ values are studied: $K = \{1000,3000,5000\}$. The detailed results are given in Figure \ref{tuning}. The top panel of Figure \ref{tuning} displays the performance of the TextCNN model. The red line in the first barplot is the prediction accuracy for the benchmark model. We find the resulting prediction
accuracy (Acc) of the reduced model increases as $K$ becomes larger, while the parameter reduction ratio (Prr) and the reduction ratio for averaged text sequence (Trr) decrease. In the case of $K=3000$, we can see the parameter reduction ratio (Prr) is more than 95\%, but there is almost no accuracy loss. This suggests that
the benchmark model can be substantially compressed with little sacrifice in predictive power. Additionally, the averaged text sequence is substantially reduced based on the dictionary screening. This indicates that there might be some redundant information in the original text that contributes less to the text classification. The bottom panel of Figure \ref{tuning} presents the results for the TextBiLSTM model, which are very similar to the findings of TextCNN.

\csubsection{Performance of Compression Results }

On the one hand, the proposed dictionary screening method aims to compress
a text classification model as much as possible. On the other hand, an over compressed model might suffer from a significant loss of prediction accuracy. Thus, it is of great importance to understand the
trade-off between prediction accuracy and model compression.
Obviously, they should be appropriately balanced. In this subsection, we report the fine-tuned compression results so that
their best performance can be demonstrated. For the best empirical performance, we try to extend the search scope for more tuning parameters. Accordingly, every value in $\{1000,2000,\cdots,10000\} $(e.g., with an interval of 1000) is tested for $K$ in this subsection. In our case, we expect that the parameter reduction ratio (Prr) will be no less than 50\% and the accuracy loss will be no more than 2\%. The best results in terms of the above criteria are summarized in Table \ref{sim1}. From Table \ref{sim1}, we can draw the following conclusions. First, for all cases, the benchmark models can be compressed substantially using the dictionary screening method with little sacrifice of accuracy. For example, the value
of Prr is more than 95\% in the case of TextCNN on Sougou News with only 0.31\% sacrifice of accuracy.
Second, we report that for
half of the cases, the prediction accuracy of the reduced model is higher
than that of the benchmark model (e.g., $\Delta$Acc is smaller than zero). For instance, for the TextBiLSTM model on DBPedia, the prediction accuracy of the baseline model
is as high as 97.77\%, while that of the reduced model is further
improved to 97.99\%. To summarize, we find that the proposed dictionary screening method works quite well on all models and datasets under consideration.

\begin{table}[htp]
\caption{Fine-tuned dictionary screening results for all model and dataset combinations with the best performance. ARP stands for the Amazon Review Polarity dataset. Benchmark Acc is the prediction accuracy of the benchmark model. Reduced Acc is the prediction accuracy of the reduced model. $\Delta$Acc is the difference between the benchmark Acc and reduced Acc. Prr is the parameter reduction ratio, Drr is the dictionary reduction ratio, and Trr is the reduction ratio for averaged text sequence. All computed values are in \% units. 
 }\label{sim1}
\vspace{0.25cm}
\renewcommand\arraystretch{1.5}
\centering
\begin{tabular}{ccccccccc}
\hline\hline
{}&{Model}&\makecell[c]{Benchmark\\Acc}&\makecell[c]{Reduced\\Acc}&{$\Delta$Acc}&{Prr} &{Drr}&{Trr}\\ \hline
\multirow{4}{*}{$TextCNN$}
&{AG's News} &89.37  &89.43 & -0.06 & 95.24 &96.81 & 34.54 \\
&{Sougou News} &95.56  &95.25 & 0.31 & 97.28 & 98.19 & 48.35 \\ 
&{DBPedia} & 98.41 & 97.88 & 0.53 & 98.97 &99.27 & 27.17\\ 
&{ARP} & 92.70 & 91.45 & 1.25 & 99.74 &99.66 &13.75\\
\hline

\multirow{4}{*}{$TextBiLSTM$}
&{AG's News} &89.58  &89.72 &-0.14 & 92.65 &94.68 & 24.25  \\
&{Sougou News} & 95.77 &95.34  & 0.43 & 96.98  & 98.19 & 47.53\\ 
&{DBPedia} & 97.77 &97.99  & -0.22 &  99.33 & 99.56&35.14\\ 
&{ARP} & 92.16 & 92.26 & -0.10 & 99.78 &99.87 &24.75\\ \hline\hline
\end{tabular}
\end{table}

\csubsection{Competing Methods }

\begin{table}[htp]
\caption{Results of the three competing methods (e.g., CPE, TF-IDF, and $t$-statistics). ARP stands for the Amazon Review Polarity dataset. Reduced Acc is the prediction accuracy of the reduced model. For each model and dataset combination, the reduced models were trained with dictionaries of equal size. Trr is the reduction ratio for averaged text sequence. The values outside of brackets are the results of the propose CPE method. The first value in brackets is the result obtained by TF-IDF, and the second value is the result by $t$-statistic. All computed values are in \% units. 
 }\label{sim2}
\vspace{0.25cm}
\renewcommand\arraystretch{1.5}
\centering
\begin{tabular}{cccc}
\hline\hline
{}&{Model}&{Reduced Acc}&{Trr}\\ \hline
\multirow{4}{*}{$TextCNN$}
&{AG's News} & 89.66 (89.33, 87.91) & 34.54 (29.19, 46.07)  \\
&{Sogou News} & 94.71 (94.66, 94.68) & 57.54 (44.39, 55.56) \\ 
&{DBPedia} & 97.89 (97.89, 17.23) & 27.17 (25.36, 93.12)  \\ 
&{ARP} & 90.73 (90.49, 89.77) & 25.50 (24.25, 32.50)  \\
\hline

\multirow{4}{*}{$TextBiLSTM$}
&{AG's News} & 90.49 (90.09, 89.89) & 18.07 (13.54, 23.84)   \\
&{Sogou News} & 95.47 (95.14, 95.31) & 47.53 (37.53, 52.54)\\ 
&{DBPedia} & 97.99 (98.00, 17.39) & 35.14 (30.80, 99.82) \\ 
&{ARP} & 92.32 (92.24, 91.55) & 24.75 (24.25, 30.00) \\ \hline\hline
\end{tabular}
\end{table}

Because the key step of the proposed dictionary screening method is to evaluate the importance of each keyword in the dictionary. We compare the performance of the proposed evaluating method, CPE, with two other competing methods. These are the TF-IDF and $t$-statistics methods, which are described in subsection 4.4. For a fair comparison, the new dictionaries constructed with the three methods are of equal size (e.g., the dictionary reduction ratio, Drr, is the same) for each model and dataset combination. As a result, only the reduced prediction accuracy (Acc) and reduction ratio for averaged text sequence (Trr) are presented in Table \ref{sim2}. From Table \ref{sim2}, we can obtain the following conclusions. First, we can see that the $t$-statistics method is not stable in evaluating the importance of keywords because its results for the DBPedia dataset were not comparable with the other methods. In the case of DBPedia, the $t$-statistic method ceases to be effective because its Trr value is nearly 100\%. This indicates that it cannot filter the important keywords from the dictionary, leading to a very low reduced Acc value. Second, in all cases, the proposed CPE method achieved a slightly higher reduced accuracy value compared with the TF-IDF method. Moreover, for most cases, by using the proposed CPE method, we can achieve a substantially reduced ratio for averaged text sequence (Trr). This indicates that the proposed CPE method can achieve a better performance in terms of predicted accuracy when keeping a relatively short text sequence. In conclusion, the proposed dictionary screening method is an efficient embedding compression method for text classification.

\csection{CONCLUSIONS}

In this paper, we propose a dictionary screening method for embedding compression in text classification tasks. The goal of this method is to evaluate the importance of each keyword in the dictionary. To this end, we develop a method called CPE to evaluate the differences in class probability estimators. With the CPE method, each keyword in the original dictionary can be screened, and only the most important keywords can be reserved. The proposed method leads
to a significant reduction in terms of parameters, average text sequence, and dictionary size. Meanwhile, the prediction power remains competitive. Extensive numerical studies are presented to demonstrate the
empirical performance of the proposed method. 

To conclude this article, we present here a number of interesting topics for future study. First, the proposed dictionary screening method involves a tuning parameter (e.g., $K$), and its optimal value for balancing prediction accuracy and parameter reduction needs to be learned. This is an important topic for future study. Second, the proposed method is only used for a text classification task. However, there are other natural language tasks, such as machine translation, question answering, and so on. The scalability of the proposed method in these tasks is also a very worthy study. Lastly, the proposed method is only conducted on English and Chinese, other language types should be investigated to test its validity.

\csection{ACKNOWLEDGEMENT}

Zhou’s research is supported in part by the National Natural Science Foundation of China (Nos. 72171226, 11971504), the Beijing Municipal Social Science Foundation (No. 19GLC052), and the National Statistical Science Research Project (No. 2020LZ38). Wang’s research is partially supported by the National Natural Science Foundation of China (No. 11831008) and the Open Research Fund of the Key Laboratory of Advanced Theory and Application in Statistics and Data Science (KLATASDS-MOE-ECNU-KLATASDS2101).

\bibliographystyle{asa}
\bibliography{DS}
\end{document}